\pdfoutput=1

\documentclass[11pt]{article}

\usepackage{acl}

\usepackage{times}
\usepackage{latexsym}

\usepackage[T1]{fontenc}

\usepackage[utf8]{inputenc}

\usepackage{microtype}

\usepackage{amsmath,amsthm,amssymb}
\usepackage{mathrsfs}
\usepackage{graphicx}
\usepackage{tabularx}
\usepackage{multirow}
\usepackage{booktabs}
\usepackage{xcolor}
\usepackage{float}
\usepackage{mathtools}

\title{Counter-GAP: Counterfactual Bias Evaluation through \\ Gendered Ambiguous Pronouns}

\author{Zhongbin Xie$^{1}$, Vid Kocijan$^{2}$, Thomas Lukasiewicz$^{3,1}$, Oana-Maria Camburu$^{4}$ \\
         $^{1}$\,University of Oxford \ $^{2}$\,Kumo.ai  \ $^{3}$\,TU Wien \\
         $^{4}$\,University College London \\
         \texttt{zhongbin.xie@cs.ox.ac.uk}, \texttt{vid@kumo.ai}, \\
         \texttt{thomas.lukasiewicz@tuwien.ac.at}, \texttt{o.camburu@ucl.ac.uk}
}

\begin{document}
\maketitle
\begin{abstract}
Bias-measuring datasets play a critical role in detecting biased behavior of language models and in evaluating progress of bias mitigation methods. In this work, we focus on evaluating gender bias through coreference resolution, where previous datasets are either hand-crafted or fail to reliably measure an explicitly defined bias. To overcome these shortcomings, we propose a novel method to collect diverse, natural, and minimally distant text pairs via counterfactual generation, and construct Counter-GAP, an annotated dataset consisting of 4008 instances grouped into 1002 quadruples. We further identify a bias cancellation problem in previous group-level metrics on Counter-GAP, and propose to use the difference between inconsistency across genders and within genders to measure bias at a quadruple level. Our results show that four pre-trained language models are significantly more inconsistent across different gender groups than within each group, and that a name-based counterfactual data augmentation method is more effective to mitigate such bias than an anonymization-based method.
\end{abstract}

\section{Introduction}
\label{intro}
It is a common practice to train state-of-the-art natural language processing (NLP) models by unsupervised pre-training and supervised fine-tuning~\citep[{e.g.},][]{devlin-etal-2019-bert,joshi-etal-2020-spanbert}, both of which rely heavily on large corpora of real-world text. However, these corpora often reflect societal stereotypes and may lead to models exhibiting biased behaviors~\citep{parrots}. Hence, much research effort has been put to reveal and mitigate unintended biases~\citep{meade-etal-2022-empirical}.  

While early work focuses on detecting and mitigating gender bias in the space of word embeddings~\citep[{e.g.},][]{bolukbasi-etal-2016-nips}, recent approaches turn to design bias-measuring datasets on specific NLP tasks~\citep{nangia-etal-2020-crows,nadeem-etal-2021-stereoset,barikeri-etal-2021-redditbias}. In this work, we focus on gender bias in coreference resolution and adopt a kind of representational harm~\citep{blodgett-etal-2020-language} to define gender fairness: a gender-neutral model should rely on the semantic information, rather than on the gender information contained in the texts, to make predictions. Otherwise, a model should be considered gender-biased. In line with this definition, WinoBias~\citep{zhao-etal-2018-gender} and WinoGender~\citep{rudinger-etal-2018-gender} leverage pairs of minimally distant sentences, {i.e.}, two sentences that contain the same semantic information but different gender information, to measure models' performance difference in resolving pronouns of different genders under the same context. This minimally distant setting enables us to isolate the influence of gender information on model predictions. 

A limitation of WinoBias and WinoGender is that they are made up of \textit{hand-crafted sentences}, which prevents us from measuring gender bias in the more diverse real-world scenarios. An alternative to overcome this shortcoming is the GAP dataset~\citep{webster-etal-2018-mind}, which exploits linguistic patterns to automatically extract instances from a \textit{real-world corpus}. 
However, since GAP's masculine and feminine instances cannot be grouped into minimally distant pairs, we are not sure whether a difference in model performance is due to different gender information or to different semantic information. For example, compared to masculine instances, GAP's feminine instances have more candidate entities serving as distractors, and longer distance between the correct name and the pronoun~\citep{vid-etal-2021-aaai}. So, it is \textit{not equally hard to resolve} the masculine and feminine instances in GAP. 
Hence, the performance difference between masculine and feminine instances on GAP is not a reliable measure of gender bias according to the above definition of gender fairness.\footnote{In GAP~\citep{webster-etal-2018-mind}, the authors did not explicitly describe the fairness definiton that they adopt.}

Given these observations, we propose a novel method to construct coreference-resolution-based bias-mea\-su\-ring datasets consisting of minimally distant text pairs that originate from real-world corpora. Specifically, we leverage the method from GAP~\citep{webster-etal-2018-mind} to extract original instances containing gendered ambiguous pronouns, and generate minimally distant instances by asking the counterfactual question ``How would the prediction change if we swapped the roles of masculine and feminine people in this context?''~\citep{garg-etal-2019-aies}. The resulting instances are grouped into quadruples, each of which consists of an original, a gender-controlled, and two gender-swapped instances. 
An example is shown in Table~\ref{tab:counter-gap}.

Furthermore, we find that bias in different directions may be canceled out if we aggregate the results by performance difference across groups of instances, and we call this problem \textit{bias cancellation}. To alleviate it, we propose a new metric, inconsistency across genders, to measure bias at the quadruple level. We also leverage the gender-controlled instances to disentangle inconsistency \textit{within} genders from inconsistency \textit{across} genders, so that we can eliminate the impact of name perturbations.

Our contributions are as follows: (\romannumeral1) We propose a novel method to construct coreference resolution datasets consisting of diverse, natural, and minimally distant instances to reliably detect gender bias. (\romannumeral2) We apply our method to online books and collect Counter-GAP, an annotated dataset with 4008 instances grouped into 1002 quadruples. (\romannumeral3)~We propose a new metric, the difference ($\Delta I$) between inconsistency across genders and within genders, to alleviate the bias cancellation problem of previous metrics. (\romannumeral4)~We use Counter-GAP to empirically evaluate four pre-trained language models and two debiasing methods based on Counterfactual Data Augmentation~\citep[CDA, ][]{zhao-etal-2018-gender,webster-etal-2020-arxiv}. Our results show that $\Delta I$ can detect significant gender bias hidden by group-level performance difference, and that name-based CDA is more effective than vanilla CDA in mitigating such bias.\footnote{The dataset and code are available at \url{https://github.com/x-zb/Counter-GAP}.}

\begin{table*}[t]
\centering
\small
\begin{tabularx}{\textwidth}{lX}
\toprule
{original}&{
\underline{\textcolor{cyan}{Tom}} did not appear to hear this, but tried to keep up the conversation with \textcolor{violet}{Julia}, desiring to have it appear that they were intimate friends; but the young lady gave brief replies, and finally, turning away, devoted herself once more to \textcolor{blue}{Herbert}, much to \underline{\textcolor{cyan}{Tom}}'s disgust. In fact, what he saw made \underline{\textcolor{cyan}{Tom}} pass a very unpleasant evening, and when, on their return home, \textcolor{orange}{Maria} suggested that \textcolor{violet}{Julia} had taken a fancy to \textcolor{blue}{Herbert}, \underline{\textbf{he}} told her to mind her own business.}\\ 
\hline
{gender-controlled}&{
\underline{\textcolor{blue}{Herbert}} did not appear to hear this, but tried to keep up the conversation with \textcolor{orange}{Maria}, desiring to have it appear that they were intimate friends; but the young lady gave brief replies, and finally, turning away, devoted herself once more to \textcolor{cyan}{Tom}, much to \underline{\textcolor{blue}{Herbert}}'s disgust. In fact, what he saw made \underline{\textcolor{blue}{Herbert}} pass a very unpleasant evening, and when, on their return home, \textcolor{violet}{Julia} suggested that \textcolor{orange}{Maria} had taken a fancy to \textcolor{cyan}{Tom}, \underline{\textbf{he}} told her to mind her own business.}\\ 
\hline
{gender-swapped-1}&{
\underline{\textcolor{orange}{Maria}} did not appear to hear this, but tried to keep up the conversation with \textcolor{blue}{Herbert}, desiring to have it appear that they were intimate friends; but the young \textit{gentleman} gave brief replies, and finally, turning away, devoted \textit{himself} once more to \textcolor{violet}{Julia}, much to \underline{\textcolor{orange}{Maria}}'s disgust. In fact, what \textit{she} saw made \underline{\textcolor{orange}{Maria}} pass a very unpleasant evening, and when, on their return home, \textcolor{cyan}{Tom} suggested that \textcolor{blue}{Herbert} had taken a fancy to \textcolor{violet}{Julia}, \underline{\textbf{she}} told \textit{him} to mind \textit{his} own business.}\\ 
\hline
{gender-swapped-2}&{
\underline{\textcolor{violet}{Julia}} did not appear to hear this, but tried to keep up the conversation with \textcolor{cyan}{Tom}, desiring to have it appear that they were intimate friends; but the young \textit{gentleman} gave brief replies, and finally, turning away, devoted \textit{himself} once more to \textcolor{orange}{Maria}, much to \underline{\textcolor{violet}{Julia}}'s disgust. In fact, what \textit{she} saw made \underline{\textcolor{violet}{Julia}} pass a very unpleasant evening, and when, on their return home, \textcolor{blue}{Herbert} suggested that \textcolor{cyan}{Tom} had taken a fancy to \textcolor{orange}{Maria}, \underline{\textbf{she}} told \textit{him} to mind \textit{his} own business.}\\ 
\bottomrule
\end{tabularx}
\caption{Counterfactual generation of a quadruple in Counter-GAP. Personal names and their genders are depicted in colors: masculine names are in \textcolor{blue}{blue} and \textcolor{cyan}{cyan}; feminine names are in \textcolor{violet}{violet} and \textcolor{orange}{orange}. The target pronoun is in \textbf{bold} and \underline{underlined}; also \underline{underlined} is the true coreferent name. Other words constitute the context, and words in \textit{italic} are gendered words swapped according to the gendered words list.}
\label{tab:counter-gap}
\end{table*}

\section{Dataset Construction}
The Counter-GAP dataset is derived from 1575 fictional books in Project Gutenberg\footnote{\url{https://www.gutenberg.org/}} and BookCorpus~\citep{zhu-etal-2015-iccv}. It is constructed through a generic multi-stage process, as described below.
Here, we follow the GAP dataset~\citep{webster-etal-2018-mind} and focus only on the English language, as well as adopting a notion of binary gender.

\subsection{Original Instance Extraction}
First, we detect all the occurrences of personal names and pronouns in a book with a dependency parser and a named entity recognizer (NER).\footnote{We use Spacy (\url{https://spacy.io/}).} For each occurrence of a gendered non-reflexive pronoun (\textit{he}, \textit{him}, \textit{his}, \textit{she}, \textit{her}, \textit{hers}), we extract a surrounding context that consists of a maximum of five sentences and contains exactly two masculine and two feminine personal names. Personal names are identified by NER tags, and the gender specification of a name is determined by statistics from a gender-guesser.\footnote{\url{https://pypi.org/project/gender-guesser/}. As we focus on English books, we use the default setting where the gender of a name is first considered according to its use in English-speaking countries.} Genders for titled names ({e.g.}, \textit{Mr.\ Smith}) are assumed from the traditional gender associations of those titles.

Second, we select the subset of contexts that contain gendered ambiguous pronouns as defined by the following three patterns from GAP~\citep{webster-etal-2018-mind} (henceforth, the gendered ambiguous pronoun is called \textit{target pronoun}, and the two names that are gender-consistent with the target pronoun are called \textit{candidate names}):
\begin{itemize}
    \item \textbf{F{\small INAL}P{\small RO}.} Both candidate names must be in the same sentence, and the target pronoun may appear in the same or directly following sentence.  
    \item \textbf{M{\small EDIAL}P{\small RO}.} The first candidate name must be in the sentence directly preceding the target pronoun and the second candidate name, both of which must be in the same sentence. The target pronoun must be in an initial subordinate clause or be a possessive in an initial prepositional phrase.
    \item\textbf{I{\small NITIAL}P{\small RO}.} Both the candidate names and the target pronoun must be in the same sentence, and the target pronoun must be in an initial subordinate clause or a possessive in an initial prepositional phrase.
\end{itemize}
After filtering, we get 2585 contexts and adopt them as original instances.

\subsection{Counterfactual Generation}
\label{sec:couter-gen}

We generate minimally distant instances in Counter-GAP through two counterfactual generation functions. An example is illustrated in Table~\ref{tab:counter-gap}. Formally, we denote an original instance as $x_o = s(P, C_1, C_2, O_1, O_2)$, where $P$ is the target pronoun to be resolved, $C_1$ and $C_2$ are the two candidate names that are gender-consistent with $P$, $O_1$ and $O_2$ are two personal names of the opposite gender, and $s(\cdot)$ denotes the context around these mentions. 

\paragraph{Gender-controlled generation.} We swap all the occurrences of $C_1$ and $C_2$, and of $O_1$ and $O_2$, to generate a gender-controlled instance $x_c = s(P, C_2, C_1, O_2, O_1)$.
We choose to swap names within an instance instead of introducing new names, so that the candidate names naturally occur in the same real-world context.

\paragraph{Gender-swapped generation.} We first substitute all gendered words with their opposite gendered words ({e.g.}, man$\rightarrow$woman, he$\rightarrow$she),\footnote{We adopt an augmented list of gendered words from \citep{zhao-etal-2018-gender}.} and swap all the occurrences of $C_1$ and $O_1$ (or $O_2$), $C_2$ and $O_2$ (or $O_1$). As a result, we obtain two gender-swapped instances $\widetilde{x_o} = \tilde{s}(\tilde{P}, O_1, O_2, C_1, C_2)$ and $\widetilde{x_c} = \tilde{s}(\tilde{P}, O_2, O_1, C_2, C_1)$, where $\tilde{s}(\cdot)$ is the context with all the gendered words substituted in $s(\cdot)$, and $\tilde{P}$ is the opposite-gendered pronoun for $P$. We call $x_o, x_c, \widetilde{x_o}, \widetilde{x_c}$ minimally distant instances, in that the words at the same position in the context ($s(\cdot)$ or $\tilde{s}(\cdot)$) are either the same (for gender-neutral words) or have the same role but opposite gender (for gendered words).

We consider a generated counterfactual instance to be invalid if (\romannumeral1) it contradicts commonsense knowledge, {e.g.}, 
historical people being of the opposite gender; or (\romannumeral2) the meaning of the counterfactual is different from the original, resulting in the gold coreference labels changing or becoming undetermined. To tackle these, we take three measures. First, we extract original instances mainly from fictional books, whose content is less likely to involve real-world people. 
Second, during human annotation (Section \ref{sec-annotation}), we explicitly ask annotators to validate whether an instance contradicts commonsense knowledge, and discard such instances. Third, we discard the whole quadruple $(x_o, x_c, \widetilde{x_o}, \widetilde{x_c})$ if not all of its four instances get the same majority labels from annotators.\footnote{Note that discarding inconsistent quadruples can also cover some error cases caused by the gender-guesser's incorrect prediction. For example, if an incorrect gender prediction occurs for a TRUE coreferent name, the target pronoun and the TRUE coreferent name will not be gender-consistent, and this may confuse the annotators, leading to inconsistent labels.} Here,  \textit{same label} means the coreferent names' positions are the same in the context (e.g., in Table~\ref{tab:counter-gap}, the position of ``Tom'' in the original instance and that of ``Maria'' in the gender-swapped-1 instance).

\subsection{Human Annotation}
\label{sec-annotation}
We use Amazon Mechanical Turk to collect coreference labels for all the $2585$ original instances and their counterfactual counterparts (hence, $2585 \times 4$ instances in total). Each instance was assigned to three annotators.
Annotation instructions and a sample task interface are presented in Appendix~\ref{appendix-ui}. Specifically, we ask annotators to highlight token spans that are coreferent with the target pronoun. We adopt majority vote to aggregate the collected annotations,
and generate a \verb|TRUE/FALSE| label for each of the two candidate names indicating if it is coreferent with the target pronoun.

After discarding quadruples containing invalid counterfactuals as discussed in Section~\ref{sec:couter-gen}, we further filter out quadruples containing real-world people to avoid grounding.
Next, we randomly downsample the remaining quadruples to balance the number of original masculine and feminine instances. The final Counter-GAP dataset consists of 1002 quadruples with an inter-annotator agreement\footnote{Average percentage of agreed annotations on each instance.} of $86.5\%$.

\section{Evaluation Metrics on Counter-GAP}

We use $X = (x_o, x_c, \widetilde{x_o}, \widetilde{x_c}) \in \mathcal{X}$ to denote a quadruple, and lowercased $x$ to denote an arbitrary instance, which could be each of $x_o, x_c, \widetilde{x_o}, \widetilde{x_c}$ from a quadruple. Given a model $f(\cdot)$, assume that $f(x) \in \{0,1\}$ indicates whether $f$'s prediction on instance $x$ is correct $(1)$ or not $(0)$.

\subsection{Bias Cancellation in Accuracy Difference}
\label{sec:cancel}
A so far commonly used metric is to directly compare the model's performance difference (or ratio) between different gender groups~\citep{webster-etal-2018-mind,sun-etal-2019-mitigating,blodgett-etal-2020-language}. For example, if we divide a test set $\mathcal{X}$ into a group of masculine instances $\mathcal{D}^{(m)}$ and a group of feminine instances $\mathcal{D}^{(f)}$ according to the gender information contained in the instances ({e.g.}, the gender of the target pronoun), gender bias can be measured by model $f$'s accuracy difference ($Acc_{Diff}$) on~$\mathcal{D}^{(m)}$ and~$\mathcal{D}^{(f)}$:
\begin{equation}
\label{equ:group}
    Acc_{Diff}\!=\!\frac{\sum_{x\in \mathcal{D}^{(m)}}f(x)}{\lvert\mathcal{D}^{(m)}\rvert}-\frac{\sum_{x\in \mathcal{D}^{(f)}}f(x)}{\lvert\mathcal{D}^{(f)}\rvert}.
\end{equation}

However, the above metric may suffer from \textit{bias cancellation} on Counter-GAP. Consider two quadruples from Counter-GAP. 
In the first, the model makes correct predictions on the two masculine instances and incorrect predictions on the two feminine ones. The model should be deemed gender-biased (towards masculine), since it makes different predictions for instances containing the same semantic information. If, in the second quadruple, the model makes reversed predictions, {i.e.}, correct on the two feminine instances and incorrect on the two masculine ones, it should also be deemed gender-biased, yet in the opposite direction towards feminine. However, the model's accuracies on the masculine and feminine groups are both $2/4=50\%$, making Eq.~\eqref{equ:group} equal to zero. In short, biases in opposite directions may be canceled out in some cases if we use Eq.~\eqref{equ:group} to aggregate them.

\subsection{Measuring Bias via Inconsistencies}
\label{sec:inconsist}

Given the bias cancellation problem of accuracy difference, we propose to measure gender bias through inconsistencies, i.e., whether a model's prediction is consistent on a pair of minimally distant instances. Specifically, we adopt two metrics, \textit{inconsistency across genders} ($I_{across}$):
\begin{equation}
    \label{equ:across}
    \begin{split}
    \frac{1}{4|\mathcal{X}|}\sum_{X\in \mathcal{X}}\Big(|f(x_o)\!-\!f(\widetilde{x_o})|+|f(x_c)\!-\!f(\widetilde{x_c})| \\
    +|f(x_o)\!-\!f(\widetilde{x_c})|+|f(x_c)\!-\!f(\widetilde{x_o})|\Big),
    \end{split}
\end{equation}
and \textit{inconsistency within genders} ($I_{within}$):
\begin{equation}
    \label{equ:within}
    \frac{1}{2|\mathcal{X}|}\sum_{X\in \mathcal{X}}{\Big(|f(x_o)\!-\!f(x_c)|+|f(\widetilde{x_o})\!-\!f(\widetilde{x_c})|\Big)}.
\end{equation}

Inconsistency across genders ($I_{across}$) measures inconsistency in instance pairs containing two instances of different genders, while inconsistency within genders ($I_{within}$) measures inconsistency in instance pairs containing two instances of the same gender. The two instances in a pair should be minimally distant ({i.e.}, from the same quadruple) to guarantee that they contain the same semantic information. Since Counter-GAP adopts personal names as proxies for person entities, we need to disentangle the part of inconsistency caused by different names ($I_{within}$) from that caused by different genders ($I_{across}$).
Therefore, our final metric to measure gender bias is 
\begin{equation}
    \label{equ:delta}
    \Delta I=I_{across}\!-\!I_{within}.
\end{equation}
In practice, a positive $\Delta I$ indicates biased behaviors of the model, while a zero or negative $\Delta I$ means that the measured inconsistency across genders are mostly noises from name perturbations, thus no bias can be detected.

\section{Bias Evaluation on Counter-GAP}

For evaluation, we adopt the coreference resolution system \verb|c2f-coref|\footnote{We use the implementations from \url{https://github.com/mandarjoshi90/coref}.}~\citep{lee-etal-2018-higher} based on four pre-trained language models: BERT-base/large and SpanBERT-base/large~\citep{joshi-etal-2020-spanbert}. 
All four models are fine-tuned on OntoNotes~\citep{pradhan-etal-2012-conll},\footnote{Since the annotation conventions of OntoNotes are a little different from those of Counter-GAP, we omitted the abbreviation period ``.'' in titles like ``Mr.'', ``Mrs.'', and ``Dr.'' in Counter-GAP during evaluation.} and training details are shown in Appendix~\ref{appendix-train}. In our evaluation, no candidate names are provided as input to the models, and models are responsible to detect candidate names in the text by themselves. A model's prediction on an instance is considered correct if the candidate name with gold label TRUE and none of those with gold label FALSE are in the target pronoun's coreferent cluster.

\subsection{Results}

Results for gender bias measured on Counter-GAP by accuracy difference (Eq.~\eqref{equ:group}) and $\Delta I$ (Eq.~\eqref{equ:delta}) are shown in Tables~\ref{tab:cancel-0} and~\ref{tab:inconsist}, respectively. In Table~\ref{tab:inconsist}, we also report the inconsistency metrics on each gender group (M, F) and each swapping direction (M2F, F2M), together with their differences.

Results in Table~\ref{tab:inconsist} show that for all four models, not only $I_{across}$ is larger than $I_{within}$ ($\Delta I$ being positive), but also the difference is statistically significant, which indicates biased behaviors in these models. Note that the absolute values of accuracy difference ($Acc_{Diff}$) in Table~\ref{tab:cancel-0} are in general smaller than the corresponding values of $\Delta I$ in Table~\ref{tab:inconsist}, and $Acc_{Diff}$ for BERT-large even becomes statistically insignificant, which is contrary to the well-known conclusion that BERT encodes social bias~\citep{nadeem-etal-2021-stereoset}. This brings evidence towards the bias cancellation problem, {i.e.}, bias measured by accuracy difference (Eq.~\eqref{equ:group}) may be canceled out compared to that measured by inconsistency difference ($\Delta I$). 

Regarding the effect of model size on gender bias, results from both metrics show that larger models seem to be less biased than smaller models. Note that both our large and base models are (pre-) trained on the same datasets, but in general larger language models are pre-trained on larger amount of data, so they are still at a higher risk of exhibiting biased behaviors~\citep{parrots}.

Regarding the detected bias direction, different metrics provide information from different perspectives. We can learn from the sign of $Acc_{{Diff}}$ in Table~\ref{tab:cancel-0} that the overall bias directions of these models are all towards masculine. In Table~\ref{tab:inconsist}, all of the Diff.\ for $I_{within}$ are negative, indicating a larger inconsistency within the feminine group. All of the Diff.\ for $I_{across}$ being negative indicates that inconsistency will increase when we change genders in an originally feminine context.

\begin{table}[h]
    \centering
    \small
    \begin{tabular}{l|ccc}
    \toprule
    {Models} & {$Acc_{M}$} & {$Acc_{F}$} & {$Acc_{Diff}$} \\
    \hline
    {BERT-base} & {63.12\%} &  {59.53\%} & {+3.59\%$^\mathbf{*}$} \\
    {BERT-large} & {72.60\%} & {72.11\%} & {+0.50\%} \\
    {SpanBERT-base} & {71.36\%} & {69.06\%} & {+2.30\%$^\mathbf{*}$} \\
    {SpanBERT-large} & {77.25\%} & {75.40\%} & {+1.85\%$^\mathbf{*}$} \\
    \bottomrule
    \end{tabular}
    \caption{Gender bias measured by Eq.~\eqref{equ:group} on Counter-GAP. We report accuracy on masculine instances ($Acc_{M}$), feminine instances ($Acc_{F}$), and their difference ($Acc_{Diff}=Acc_{M}-Acc_{F}$). A ``$\mathbf{*}$'' means that the difference is statistically significant ($p<0.01$) under one-sided bootstrap resampling~\citep{graham-etal-2014-randomized}.}
    \label{tab:cancel-0}
\end{table}

\begin{table*}[t]
    \centering
    \small
    \begin{tabular}{p{0.135\textwidth}|p{0.05\textwidth}<{\centering}p{0.05\textwidth}<{\centering}p{0.06\textwidth}<{\centering}p{0.05\textwidth}<{\centering}|p{0.05\textwidth}<{\centering}p{0.05\textwidth}<{\centering}p{0.06\textwidth}<{\centering}p{0.05\textwidth}<{\centering}|p{0.14\textwidth}<{\centering}}
    \toprule
    \multirow{2}*{Models}&\multicolumn{4}{c|}{inconsistency within genders}&\multicolumn{4}{c|}{inconsistency across genders}&{$\Delta I\!=\!I_{across}$} \\
    \cline{2-9}{}&{M}&{F}&{Diff.}&{$I_{within}$}&{M2F}&{F2M}&{Diff.}&{$I_{across}$}&{$-\!I_{within}$} \\
    \hline
    {BERT-base} & {15.47\%} & {16.47\%} & {-1.00\%} & {15.97\%} &  {18.26\%} & {23.25\%} & {-4.99\%} & {20.76\%} & {+4.79\%$^\mathbf{*}$} \\
    {BERT-large} & {10.28\%} & {10.28\%} & {0.00\%} & {10.28\%} & {10.88\%} & {14.27\%} & {-3.39\%} & {12.57\%} & {+2.30\%$^\mathbf{*}$} \\
    {SpanBERT-base} & {9.98\%} & {12.18\%} & {-2.20\%} & {11.08\%} & {12.18\%} & {15.07\%} & {-2.89\%} & {13.62\%} & {+2.54\%$^\mathbf{*}$} \\
    {SpanBERT-large} & {5.79\%} & {6.29\%} & {-0.50\%} & {6.04\%} & {6.89\%} & {8.18\%} & {-1.30\%} & {7.53\%} & {+1.50\%$^\mathbf{*}$} \\
    \bottomrule
    \end{tabular}
    \caption{Gender bias measured by the inconsistency metrics on Counter-GAP. We report inconsistency within genders on masculine instances (M), feminine instances (F), and their difference (Diff. = M - F), as well as inconsistency within genders on all the instances ($I_{within}$). We also report inconsistency across genders on quadruples generated by transforming masculine instances to feminine instances (M2F), transforming feminine instances to masculine instances (F2M), and their difference (Diff. = M2F - F2M), as well as inconsistency across genders on all the quadruples ($I_{across}$). $\Delta I=I_{across}\!-\!I_{within}$ measures gender bias, where a ``$\mathbf{*}$'' means that the difference is statistically significant ($p<0.01$) under one-sided bootstrap resampling~\citep{graham-etal-2014-randomized}.}
    \label{tab:inconsist}
\end{table*}

\subsection{No Bias Between the Original and Counterfactual Instances}
\label{sec:systematic bias}
Since the gender-swapped instances in Counter-GAP are generated automatically, although they have been validated by annotators, we still check whether there is a systematic bias towards the original or counterfactual instances. To investigate this, we measure two statistics. First, we measure a model's accuracy on instances with the original gender ($x_o, x_c$) and the counterfactual gender ($\widetilde{x_o}, \widetilde{x_c}$), and report their difference. From Table~\ref{tab:system}, we see that the differences are very small and not statistically significant. Second, we measure the correlation between the inconsistency across genders score ($|f(x_o)\!-\!f(\widetilde{x_o})|+|f(x_c)\!-\!f(\widetilde{x_c})|+|f(x_o)\!-\!f(\widetilde{x_c})|+|f(x_c)\!-\!f(\widetilde{x_o})|$) and the original gender of a quadruple $X$. From Table~\ref{tab:system}, we see that the values of Spearman's $\rho$ are all close to zero, indicating no significant correlations. Hence, we conclude that the counterfactual instances in Counter-GAP do not introduce systematic bias.

\begin{table}
    \centering
    \small
    \begin{tabular}{p{0.14\textwidth}|p{0.04\textwidth}<{\centering}p{0.05\textwidth}<{\centering}p{0.058\textwidth}<{\centering}|p{0.055\textwidth}<{\centering}}
    \toprule
    \multirow{2}*{Models}&\multicolumn{3}{c|}{Accuracy}&{Spear-}\\
    \cline{2-4}{}&{Orig.}&{Counter.}&{Diff.}&{man's$\rho$}  \\
    \hline
    {BERT-base} & {61.58\%}&{61.08\%}&{+0.50\%}&{-0.083} \\
    {BERT-large} & {72.06\%}&{72.65\%}&{-0.60\%}&{-0.065} \\
    {SpanBERT-base} & {70.21\%}&{70.21\%}&{0.00\%}&{-0.060} \\
    {SpanBERT-large} & {76.55\%}&{76.10\%}&{+0.45\%}&{-0.030} \\
    \bottomrule
    \end{tabular}
    \caption{Results on systematic bias evaluation. We report accuracy on instances with the original gender (Orig.), with the counterfactual gender (Counter.), and their difference (Diff.= Orig. - Counter.). All the differences are not statistically significant ($p>0.01$) under one-sided bootstrap resampling~\citep{graham-etal-2014-randomized}. We also report Spearman's $\rho$ between inconsistency across genders and the original gender of a quadruple.}
    \label{tab:system}
\end{table}

\subsection{Comparison with GAP}

\begin{table*}
    \centering
    \small
    \begin{tabular}{l|cccc|c}
    \toprule
    \multirow{2}*{Datasets}&\multicolumn{4}{c|}{Accuracy}&{$\Delta I$}\\
    \cline{2-5}{}&{$Acc_{M}$}&{$Acc_{F}$}&{$Acc_{Diff}$}&{Overall}&{}  \\
    \hline
    {original GAP} & {85.10\%}&{79.80\%}&{+5.30\%}&{82.45\%} & {-- --} \\
    {our-GAP} & {75.25\%}&{78.44\%}&{-3.19\%}&{76.85\%} & {-- --} \\
    {Counter-GAP} & {77.25\%}&{75.40\%}&{+1.85\%}&{76.32\%} & {+1.50\%} \\
    \bottomrule
    \end{tabular}
    \caption{Results from SpanBERT-large on three datasets.}
    \label{tab:comparison}
\end{table*}

We further compare Counter-GAP with two GAP-like datasets: the original GAP test set~\citep{webster-etal-2018-mind} and a subset of Counter-GAP where only the original instances $x_o$ are kept (we call this dataset our-GAP). The results of SpanBERT-large on the above datasets are shown in Table~\ref{tab:comparison}. We see that the accuracy differences between masculine and feminine instances are much smaller on Counter-GAP than on the original GAP and our-GAP. This empirically verifies that datasets without minimally distant instances cannot reliably measure bias (they amplify bias in this case) due to the different semantic information contained in its masculine and feminine instances. 
Moreover, the overall direction of detected gender bias (the sign of $Acc_{Diff}$) is different on the original GAP and our-GAP, which shows that different source corpora (Wikipedia for GAP vs.\ fictions for our-GAP) may detect different bias in the model. This highlights the importance of domain diversity when using data-centric methods for bias detection.

\section{Bias Mitigation}

We evaluate two debiasing methods based on counterfactual data augmentation (CDA)~\citep{zhao-etal-2018-gender,webster-etal-2020-arxiv}: (\romannumeral1) anonymization-based CDA (a-CDA), where the training set (OntoNotes) is augmented by substituting all gendered words with their opposite gendered words, while the gold coreference labels are kept unchanged. Personal names in the training set are anonymized using place holders such as ``E1, E2, \dots''; (\romannumeral2) name-based CDA (n-CDA), where, in addition to the substitution between gendered words, masculine and feminine names also substitute each other according to their frequencies~\citep{hall-maudslay-etal-2019-name}. See Appendix~\ref{appendix-train} for more details. Performance on bias mitigation is measured by $Acc_{Diff}$ and $\Delta I$, while performance on coreference resolution is measured by overall accuracy on Counter-GAP and $F_1$ score on OntoNotes' dev set.

\begin{table*}[t]
\centering
\small
\begin{tabular}{p{0.12\textwidth}|p{0.09\textwidth}||p{0.14\textwidth}<{\centering}|p{0.14\textwidth}<{\centering}|p{0.11\textwidth}<{\centering}||p{0.11\textwidth}<{\centering}}
\toprule
{\small{Models}}&{\small{Debiasing Method}}&{$Acc_{Diff}=Acc_{M}-Acc_{F}$}&{$\Delta I=I_{across}-I_{within}$}&{\small{Overall Accuracy}}&{\small{$F_1$ on OntoNotes}}\\
\hline
\specialrule{0em}{1.5pt}{1.5pt}
\hline
\multirow{3}*{}&{none}&{+3.59\%$^*$}&{+4.79\%$^*$}&{61.33\%}&{74.39\%}\\
{BERT-base}&{a-CDA}&{+1.90\%}&{+2.30\%$^*$}&{65.17\%}&{73.91\%}\\
{}&{n-CDA}&{\textbf{+0.20\%}}&{+1.85\%$^*$}&{66.82\%}&{73.60\%}\\
\hline
\multirow{3}*{}&{none}&{+0.50\%}&{+2.30\%$^*$}&{72.36\%}&{77.35\%}\\
{BERT-large}&{a-CDA}&{+2.99\%$^*$}&{+1.75\%$^*$}&{72.95\%}&{76.96\%}\\
{}&{n-CDA}&{+0.95\%}&{+1.30\%}&{73.53\%}&{77.13\%}\\
\hline
\multirow{3}*{}&{none}&{+2.30\%$^*$}&{+2.54\%$^*$}&{70.21\%}&{77.71\%}\\
{SpanBERT}&{a-CDA}&{+5.14\%$^*$}&{+2.54\%$^*$}&{69.49\%}&{78.04\%}\\
{-base}&{n-CDA}&{+0.95\%}&{+1.25\%}&{71.03\%}&{77.70\%}\\
\hline
\multirow{3}*{}&{none}&{+1.85\%$^*$}&{+1.50\%$^*$}&{76.32\%}&{80.06\%}\\
{SpanBERT}&{a-CDA}&{+0.35\%}&{+1.45\%$^*$}&{76.72\%}&{\textbf{80.07\%}}\\
{-large}&{n-CDA}&{+0.65\%}&{\textbf{+0.15\%}}&{\textbf{77.92\%}}&{79.93\%}\\
\bottomrule
\end{tabular}
\caption{Bias mitigation results. 
For $Acc_{Diff}$ and $\Delta I$, lower is better; for overall accuracy and $F_1$ on OntoNotes, higher is better. Best results are in bold. A ``*'' on $\Delta I$ indicates that the difference is statistically significant ($p<0.01$) under one-sided bootstrap resampling~\citep{graham-etal-2014-randomized}.}
\label{tab:cda}
\end{table*}

Results are shown in Table~\ref{tab:cda}. In terms of $\Delta I$, both a-CDA and n-CDA can effectively reduce gender bias, while n-CDA is more effective than a-CDA in that its $\Delta I$ values are smaller and less significant. Comparing the results of $Acc_{Diff}$ and $\Delta I$, we discover that bias measured by $Acc_{Diff}$ tends to be more easily mitigated by the debiased methods. For example, a-CDA fails to reduce $\Delta I$ to an insignificant level for all the four models, but it succeeds to do so for $Acc_{Diff}$ on BERT-base and SpanBERT-large; n-CDA can reduce BERT-base's $Acc_{Diff}$ to an insignificant level, but fails to do so under the measurement of $\Delta I$.

Regarding the trade-off between bias mitigation and overall performance, both a-CDA and n-CDA can maintain or even increase the overall accuracy on Counter-GAP. This indicates that they do not sacrifice model performance for fairness, which is a favorable characteristic of debiasing methods. However, n-CDA achieves decreased $F_1$ scores on OntoNotes for all the four models, indicating that it is more suitable for tasks involving mostly personal names.

\section{Qualitative Analysis}

\begin{table*}[!h]
\centering
\small
\begin{tabularx}{\textwidth}{p{0.08\textwidth}X}
\toprule
{Example}&{1}\\
\hline
{original\newline\textcolor{red}{(correct)}}&
{"In fact, \textcolor{violet}{Roxanne} told me that she had scheduled an interview with a source tonight."  \textcolor{orange}{Denise} sipped at her lemonade through her straw until she found the bottom of her glass at last. \colorbox{yellow}{\underline{\textcolor{blue}{Scotty}}} told \textcolor{cyan}{Chris} that \textcolor{orange}{Denise} didn't give \colorbox{yellow}{\underline{\textbf{him}}} any particulars about why she needed to hire a private detective when she sought \colorbox{yellow}{his} advice.}\\ 
\hline
{gender-controlled\newline\textcolor{red}{(correct)}}&
{"In fact, \textcolor{orange}{Denise} told me that she had scheduled an interview with a source tonight."  \textcolor{violet}{Roxanne} sipped at her lemonade through her straw until she found the bottom of her glass at last. \colorbox{yellow}{\underline{\textcolor{cyan}{Chris}}} told \textcolor{blue}{Scotty} that \textcolor{violet}{Roxanne} didn't give \colorbox{yellow}{\underline{\textbf{him}}} any particulars about why she needed to hire a private detective when she sought \colorbox{yellow}{his} advice.} \\
\hline
{gender-swapped-1\newline\textcolor{red}{(incorrect)}}&
{"In fact, \textcolor{cyan}{Chris} told me that he had scheduled an interview with a source tonight."  \textcolor{blue}{Scotty} sipped at his lemonade through his straw until he found the bottom of his glass at last. \colorbox{yellow}{\underline{\textcolor{orange}{Denise}}} told \colorbox{yellow}{\textcolor{violet}{Roxanne}} that \textcolor{blue}{Scotty} didn't give  \colorbox{yellow}{\underline{\textbf{her}}} any particulars about why he needed to hire a private detective when he sought \colorbox{yellow}{her} advice.} \\
\hline
{gender-swapped-2\newline\textcolor{red}{(incorrect)}}&
{"In fact, \textcolor{blue}{Scotty} told me that he had scheduled an interview with a source tonight."  \textcolor{cyan}{Chris} sipped at his lemonade through his straw until he found the bottom of his glass at last. \underline{\textcolor{violet}{Roxanne}} told \colorbox{yellow}{\textcolor{orange}{Denise}} that \textcolor{cyan}{Chris} didn't give \colorbox{yellow}{\underline{\textbf{her}}} any particulars about why he needed to hire a private detective when he sought \colorbox{yellow}{her} advice.} \\
\hline
\specialrule{0em}{1.5pt}{1.5pt}
\hline
{Example}&{2}\\
\hline
{original\newline\textcolor{red}{(correct)}}&
{\textcolor{violet}{Dora} said, "You ought not to bet, especially on Sunday," and \textcolor{orange}{Alice} altered it to "You may be sure."  "Well, but what then?" \colorbox{yellow}{\underline{\textcolor{blue}{Oswald}}} asked \textcolor{cyan}{Denny}. "Out with it," for \colorbox{yellow}{\underline{\textbf{he}}} saw that \colorbox{yellow}{his} youthful friend had got an idea and couldn't get it out.}\\ 
\hline
{gender-controlled\newline\textcolor{red}{(correct)}}&
{\textcolor{orange}{Alice} said, "You ought not to bet, especially on Sunday," and \textcolor{violet}{Dora} altered it to "You may be sure."  "Well, but what then?" \colorbox{yellow}{\underline{\textcolor{cyan}{Denny}}} asked \textcolor{blue}{Oswald}. "Out with it," for \colorbox{yellow}{\underline{\textbf{he}}} saw that \colorbox{yellow}{his} youthful friend had got an idea and couldn't get it out.} \\
\hline
{gender-swapped-1\newline\textcolor{red}{(incorrect)}}&
{\textcolor{blue}{Oswald} said, "You ought not to bet, especially on Sunday," and \textcolor{cyan}{Denny} altered it to "You may be sure."  "Well, but what then?" \underline{\textcolor{violet}{Dora}} asked \colorbox{yellow}{\textcolor{orange}{Alice}}. "Out with it," for \colorbox{yellow}{\underline{\textbf{she}}} saw that \colorbox{yellow}{her youthful friend} had got an idea and couldn't get it out.} \\
\hline
{gender-swapped-2\newline\textcolor{red}{(correct)}}&
{\textcolor{cyan}{Denny} said, ""You ought not to bet, especially on Sunday,"" and \textcolor{blue}{Oswald} altered it to ""You may be sure.""  ""Well, but what then?"" \colorbox{yellow}{\underline{\textcolor{orange}{Alice}}} asked \textcolor{violet}{Dora}. ""Out with it,"" for \colorbox{yellow}{\underline{\textbf{she}}} saw that \colorbox{yellow}{her} youthful friend had got an idea and couldn't get it out.} \\
\bottomrule
\end{tabularx}
\caption{Examples from Counter-GAP. In each instance, the predicted coreference cluster from the model is highlighted in \colorbox{yellow}{yellow}, and the \textcolor{red}{correctness} of the prediction is annotated in the first column. The target pronoun is in \textbf{bold} and \underline{underlined}; also \underline{underlined} is the true coreferent name. Other notations follow those in Table~\ref{tab:counter-gap}.}
\label{tab:case}
\end{table*}

In Table~\ref{tab:case}, we show some Counter-GAP examples with predictions from SpanBERT-large.
In Example 1, SpanBERT-large makes correct decisions for both the original and gender-controlled instance. But for the two gender-swapped instances, it either refers the target pronoun to the incorrect feminine person (Denise), or includes both feminine names (Roxanne and Denise) in the coreferent cluster, which indicates a worse performance on resolving feminine pronouns under the same context. This kind of gender bias can only be detected when we counterfactually augment the original instance. 

In Example 2, SpanBERT-large correctly finds ``Alice'' in the gender-swapped-2 instance, but confuses ``Alice'' and ``Dora'' in the gender-swapped-1 instance. This illustrates the inconsistency brought by name perturbations within the same gender, and we take this into account by subtracting inconsistency within genders from inconsistency across genders ($\Delta I=I_{across}-I_{within}$).

\section{Related Work}
\paragraph{Measuring Bias in NLP models.} Human-like biases are first detected and measured in word embeddings~\citep{bolukbasi-etal-2016-nips,caliskan-etal-2017-science,pnas,gonen-goldberg-2019-lipstick,manzini-etal-2019-black}. For pre-trained language models, \citet{may-etal-2019-measuring} adopt bleached sentence templates to contextualize target words, while most recent works leverage crowd-sourced benchmark datasets on NLP tasks such as language modeling~\citep{nangia-etal-2020-crows,nadeem-etal-2021-stereoset}, sentiment analysis~\citep{kiritchenko-mohammad-2018-examining}, dialog generation~\citep{barikeri-etal-2021-redditbias}, natural language inference~\citep{dev-etal-2020-aaai}, and machine translation~\citep{stanovsky-etal-2019-evaluating}. Our work follows GAP~\citep{webster-etal-2018-mind}, WinoBias~\citep{zhao-etal-2018-gender}, and WinoGender~\citep{rudinger-etal-2018-gender} to measure gender bias in coreference resolution, with a specific focus on collecting diverse, natural, and minimally distant instances.

\paragraph{Counterfactual Bias Evaluation.} \citet{kusner-etal-2017-nips} propose the notion of \textit{counterfactual fairness}, which requires similar predictions before and after counterfactual interventions in casual graphs. \citet{garg-etal-2019-aies} apply this notion to text classification and propose the metric of Counterfactual Token Fairness, which is similar to our inconsistency metrics, but we further distinguish inconsistency within genders from inconsistency across genders in our quadruple setting. Counterfactual Data Augmentation (CDA)~~\citep{webster-etal-2020-arxiv,zmigrod-etal-2019-counterfactual} is a widely adopted method for bias evaluation~\citep{cao-etal-2020-towards,zhang-etal-2021-double}, and we additionally focus on personal names during counterfactual generation of coreference resolution instances.

\paragraph{Name Artifacts in NLP Models.} Since neural language models do not treat personal names as interchangeable, there are various biases in the learned representations of personal names~\citep{shwartz-etal-2020-grounded,prabhakaran-etal-2019-perturbation,wolfe-caliskan-2021-low,wang-etal-2022-measuring}. Counter-GAP considers name biases as the source of gender bias, and exhibits these name biases through the task of coreference resolution.

\section{Summary and Outlook}
In this work, we proposed a method to construct minimally distant bias-measuring datasets for coreference resolution, and exemplified it in the collection of Counter-GAP. We proposed the inconsistency metric $\Delta I$ to overcome the bias cancellation problem and noise from name perturbations. We showed that four pre-trained language models exhibit significant gender bias, and name-based CDA is most effective in mitigating the detected bias.

Limitations of Counter-GAP include that around half of the instances are from historical fictions in Project Gutenburg, making the dataset less representative of contemporary bias; the rules in our method for constructing Counter-GAP are specific for English, and might not be easily adapted to languages with more complex morphology; while we recognize that gender is non-binary, we adopt the simplifying setup of binary gender construct, which prevents us from detecting gender bias against minority groups with non-binary genders. 

In future work, we will apply our method to different domains and more contemporary corpora such as news articles. Leveraging the data augmentation method for languages with grammatical genders~\citep{zmigrod-etal-2019-counterfactual}, as well as linguistic resources for non-binary genders~\citep{cao-etal-2020-towards} is also an important future direction to construct more gender- and language-inclusive datasets.

\section*{Limitations}
\label{limitations}
Counter-GAP adopts the setup of a binary gender construct, which restricts it from detecting bias against non-binary gender groups. 
Future work may extend Counter-GAP using non-binary gendered word lists, and correspondingly extend our metric (inconsistency across and within binary gender groups) for multiple gender groups.

Our method relies on specific characteristics of the English language. Directly applying it to other languages may be non-trivial. For example, languages like French or Italian adopt grammatical genders that need extra rules in our counterfactual generation method, while Chinese names are, in principle, gender neutral, which makes it impossible to identify genders from personal names. Therefore, adaptation efforts are required for researchers working on multilingual problems.  

Like many other bias-measuring datasets, Coun\-ter-GAP only serves as a diagnostic dataset. This means that, if our dataset and metric detect significant bias, we could deem a model biased; but if little or no bias is detected, we cannot guarantee that the model is unbiased. Practitioners may adopt diverse bias benchmarks before reaching a conclusion.

\section*{Acknowledgements}
During this work, Oana-Maria Camburu was supported by an Early Career Leverhulme Fellowship.
This work was also supported by the Alan Turing Institute under the EPSRC grant EP/N510129/1 and the AXA
Research Fund.
We also acknowledge the use of Oxford’s ARC facility, of the EPSRC-funded Tier 2 facilities JADE (EP/P020275/1), and of GPU computing support by Scan Computers International Ltd. 

\bibliography{anthology,custom}

\clearpage
\appendix

\section{Training Details}
\label{appendix-train}

Training details follow those by \citet{joshi-etal-2019-bert,joshi-etal-2020-spanbert}, and the hyperparameters adopted for each model during fine-tuning on OntoNotes are shown in Table~\ref{tab:hyperparam}. Specifically, each document in OntoNotes is divided into non-overlapping segments of length \verb|max_segment_len|. The segments are then encoded independently by BERT/SpanBERT to contextualized word embeddings and fed to the \verb|c2f-coref| model~\citep{lee-etal-2018-higher}. The models are fine-tuned for 20 epochs with a dropout rate of 0.3, \verb|bert_learning_rate| on parameters in BERT/SpanBERT, and \verb|task_learning_rate| on parameters in \verb|c2f-coref|. The learning rates are linearly decayed. A batch size of one document is used, where each document is randomly truncated to contain \verb|max_training_sentences| segments. All the experiments are conducted on one Tesla-V100 GPU with 32 GB memory. 

\begin{table*}
    \centering
    \small
    \begin{tabular}{|l|rrrr|}
    \hline
    {} & {BERT-base} & {BERT-large} & {SpanBERT-base} & {SpanBERT-large} \\
    \hline
    {$bert\_learning\_rate$} & {1e-5} & {1e-5} & {2e-5} & {1e-5} \\
    {$task\_learning\_rate$} & {2e-4} & {2e-4} & {1e-4} & {3e-4} \\
    {$max\_segment\_len$} & {128} & {384} & {384} & {512}  \\
    {$max\_training\_sentences$} & {11} & {3} & {3} & {3} \\
    \hline
    \end{tabular}
    \caption{Hyperparameters for fine-tuning.}
    \label{tab:hyperparam}
\end{table*}

For the debiasing method n-CDA, we adjust the bipartite graph matching method from \citet{hall-maudslay-etal-2019-name} to fit the name list in our gender-guesser. Specifically, in our name list, each first name is assigned a label in \{``male'', ``mostly male'', ``female'', ``mostly female''\} indicating its gender specification, as well as a 55-dimensional frequency vector. The value in each dimension is an integer in $[0, 13]$ that indicates the name's relative frequency in one of the 55 countries. Below, we only describe how we match ``male''  with ``female'' names; the method to match ``mostly male''  with ``mostly female'' names is the same. We build a bipartite graph where ``male'' and ``female'' names are nodes in distinct parts, and define the weight of an edge between a ``male'' and a ``female'' name as $w_{i,j} = \Vert v_i-v_j\Vert_2 \cdot (\alpha-cos\langle v_i,v_j\rangle)$, where $v_i$ and $v_j$ are the frequency vectors of the two names, and $\alpha>1$ is a hyperparameter balancing the $\ell_2$  and the cosine distance. Our motivation is to encourage a rare name to be matched with even a popular name in the same country other than another rare name in a different country, so we choose $\alpha=12/11$. Finally, we leverage a minimum weight full matching algorithm~\citep{kuhn-1955-nav} to compute the matches between the names in the two parts.

\section{Amazon Mechanical Turk Annotation Details}
\label{appendix-ui}
\begin{figure*}[htbp]
\small
\centering
\centerline{\includegraphics[width=0.99\textwidth]{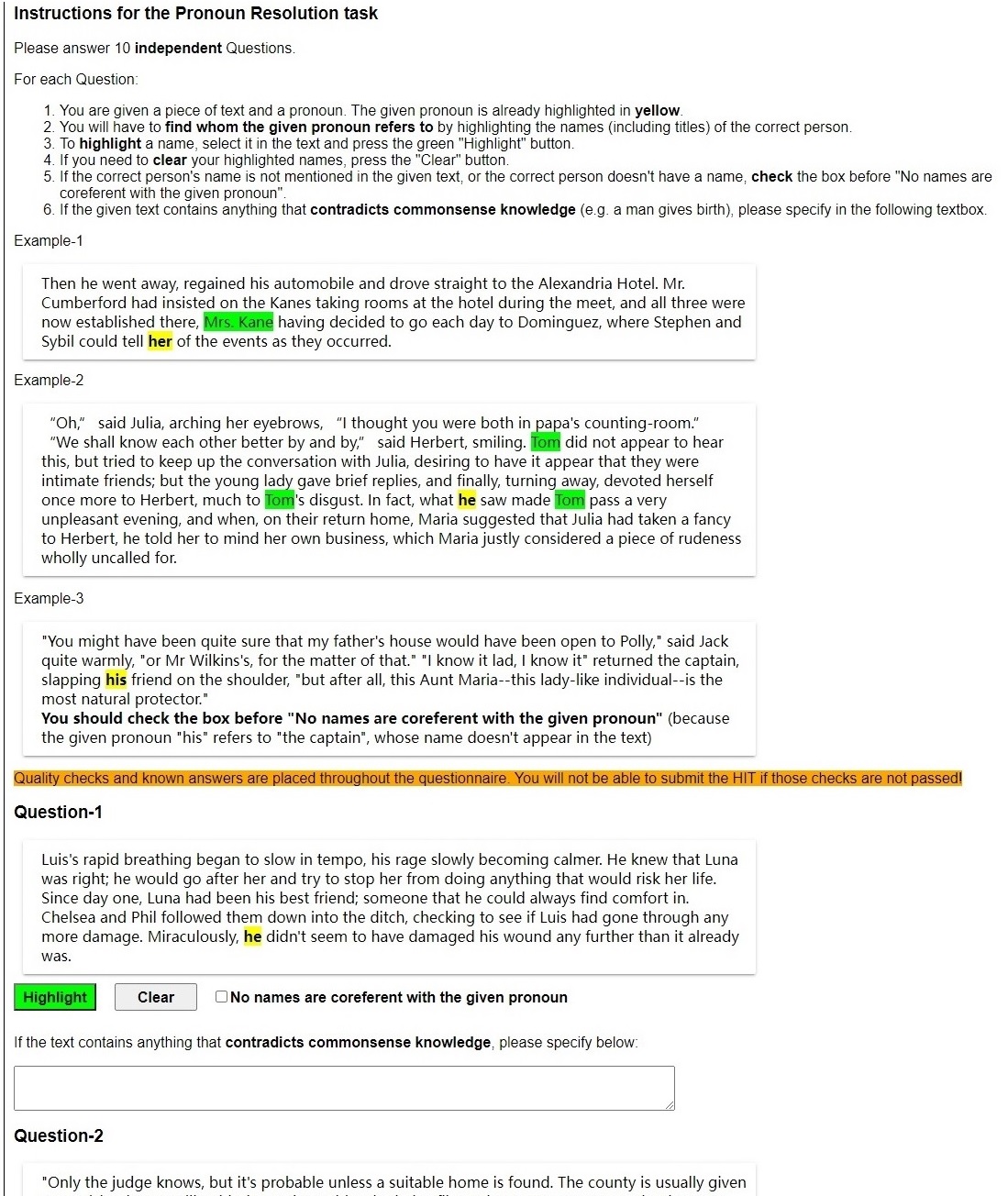}}
\caption{A sample HIT interface and annotation instructions.}
\label{fig:amt}
\end{figure*}

Our annotation instructions and a sample Human Intelligence Task (HIT) interface are shown in Figure~\ref{fig:amt}. To ensure the annotation quality, we implement a series of on-submission checks including checks on whether the selected span is a personal name, whether multiple entities are selected, whether the ``no names are coreferent'' box is misused, and so on. We require annotators to have at least a 95\% approval rate with more than 50 approved HITs. 
The average time an annotator spent on one HIT is around 30 minutes.

\section{Illustration of the Inconsistency Metrics}

\begin{figure}[h]
\centering
\centerline{\includegraphics[width=0.60\columnwidth]{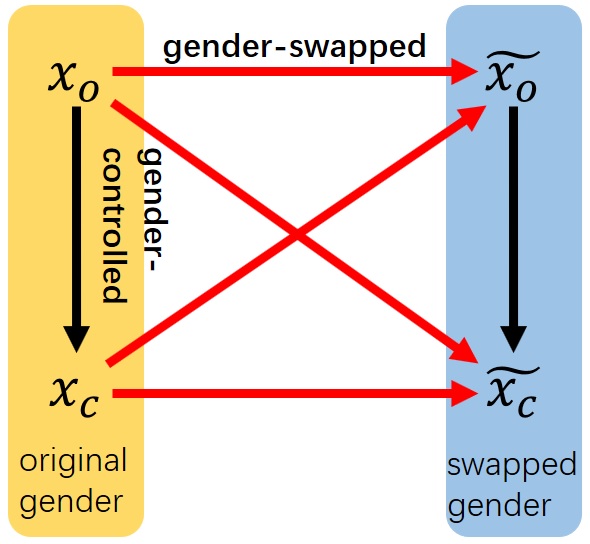}}
\caption{Illustration of the two inconsistency metrics. When computing inconsistency within genders, we use instance pairs linked by the two black arrows; when computing inconsistency across genders, we use instance pairs linked by the four red arrows.}
\label{fig:swap}
\end{figure}

A conceptual illustration of the proposed inconsistency metrics is shown in Figure~\ref{fig:swap}: inconsistency across genders ($I_{across}$) measures inconsistency in instance pairs containing two instances of different genders, while inconsistency within genders ($I_{within}$) measures inconsistency in instance pairs containing two instances of the same gender.

\end{document}